\documentclass[letterpaper]{article} 
\usepackage{aaai25}  
\usepackage{times}  
\usepackage{helvet}  
\usepackage{courier}  
\usepackage[hyphens]{url}  
\usepackage{graphicx} 
\usepackage{natbib}  
\usepackage{caption} 
\usepackage{algorithm}
\usepackage{algorithmic}
\usepackage{amsmath,amsfonts,amssymb}
\usepackage{booktabs}
\usepackage{multirow}
\usepackage{newfloat}
\usepackage{listings}
\DeclareCaptionStyle{ruled}{labelfont=normalfont,labelsep=colon,strut=off} 
\floatstyle{ruled}
\newfloat{listing}{tb}{lst}{}
\floatname{listing}{Listing}
\title{Data-Free Universal Attack \\ by Exploiting the Intrinsic Vulnerability of Deep Models}
\author{
    YangTian Yan,
    Jinyu Tian\thanks{The corresponding author.}\\
}
\affiliations{
    Faculty of Innovation Engineering, Macau University of Science and Technology\\

    naygnahz1128@gmail.com, jytian@must.edu.mo
}

\usepackage{bibentry}

\begin{document}

\maketitle

\begin{abstract}
Deep neural networks (DNNs) are susceptible to Universal Adversarial Perturbations (UAPs), which are instance-agnostic perturbations that can deceive a target model across a wide range of samples. Unlike instance-specific adversarial examples, UAPs present a greater challenge as they must generalize across different samples and models. Generating UAPs typically requires access to numerous examples, which is a strong assumption in real-world tasks. In this paper, we propose a novel data-free method called \textbf{Intri}nsic \textbf{UAP} (\textbf{IntriUAP}), by exploiting the intrinsic vulnerabilities of deep models. We analyze a series of popular deep models composed of linear and nonlinear layers with a Lipschitz constant of 1, revealing that the vulnerability of these models is predominantly influenced by their linear components. Based on this observation, we leverage the ill-conditioned nature of the linear components by aligning the UAP with the right singular vectors corresponding to the maximum singular value of each linear layer. Remarkably, our method achieves highly competitive performance in attacking popular image classification deep models without using any image samples. We also evaluate the black-box attack performance of our method, showing that it matches the state-of-the-art baseline for data-free methods on models that conform to our theoretical framework. Beyond the data-free assumption, IntriUAP also operates under a weaker assumption, where the adversary only can access a few of the victim model's layers. Experiments demonstrate that the attack success rate decreases by only $4\%$ when the adversary has access to just $50\%$ of the linear layers in the victim model. 
\end{abstract}

%
\begin{links}
    \link{Code}{https://github.com/yyt0718/Intri_Attack}
\end{links}

\section{Introduction}

Deep Neural Networks (DNNs) have shown remarkable performance in tasks such as computer vision \cite{krizhevsky2012imagenet,simonyan2014very,he2016deep}, natural language processing \cite{bahdanau2014neural,vaswani2017attention,devlin2018bert,biderman2023pythia}, etc. 
However, recent research~\cite{goodfellow2015explaining,moosavi2016deepfool,madry2019deep} has highlighted their vulnerability to adversarial attacks. Adversarial scenarios arise when models face inputs specifically designed to deceive or mislead them, often resulting in erroneous outputs. This vulnerability is not merely academic but poses significant practical challenges, especially in critical applications such as autonomous driving and facial recognition systems. As models become more integral to our daily lives, ensuring their reliability against such adversarial attacks is crucial. This motivates ongoing research to understand and mitigate these threats, aiming to develop more reliable and secure DNNs. 

Among the spectrum of adversarial strategies, \textbf{U}niversal \textbf{A}dversarial \textbf{P}erturbations (UAPs) represent a particularly important form. UAPs are image-agnostic perturbations, which means a single perturbation, once computed, can be applied to a wide variety of images to consistently mislead a machine learning model. Most of the current work on UAPs requires massive samples \cite{moosavi2017universal,mopuri2018nag,poursaeed2018generative,liu2023enhancing}. These methods, although efficient, necessitate extensive samples and significant time for iterative training and adjustment of perturbations. They also assume unrestricted access to the original training data or sizable representative datasets. This requirement limits their practicality in scenarios where data privacy or accessibility is constrained. In contrast, a few existing approaches attempt to generate UAPs without direct reliance on extensive large datasets 
\cite{mopuri2017fast,mopuri2018ask,mopuri2018generalizable,liu2019universal,zhang2020understanding,zhang2021data}. They generally follow two lines. The first involves designing batches of proxy datasets, such as using GANs to generate some proxy samples \cite{mopuri2018ask}. Another type of UAPs leverages the characteristic of the model, for instance, exploiting the observation that adversarial samples can occupy a large portion of the label space \cite{zhang2020understanding,zhang2021data}.

Recent studies have shown that specific patterns \cite{mopuri2018generalizable,9008259} or even random noise \cite{fawzi2016robustnessclassifiersadversarialrandom} can successfully attack deep models. This suggests that the existence of UAPs is primarily due to intrinsic flaws within the models themselves, rather than being learned from a large number of samples. Several studies support this view by analyzing the geometric properties of decision boundaries~\cite{goodfellow2015explaining,zhao2024detecting,su2024model,tao20233dhacker}. Along this line, in this paper, we leverage the inherent vulnerabilities of deep models to design UAPs, which we refer to as \textbf{Intri}nsic \textbf{UAP} \textbf{(IntriUAP)}, without relying on training data. Specifically, we conduct an in-depth analysis of a popular type of deep model that we call the \textbf{L}inear and \textbf{1}-\textbf{L}ipschitz \textbf{O}perator \textbf{S}ystem \textbf{(L1LOS)} and propose a novel data-free UAP method based on its intrinsic vulnerabilities. The L1LOS is composed of linear operators (layers in deep models) and nonlinear operators with a Lipschitz constant of 1. Several popular deep models, such as VGG ~\cite{simonyan2014very}, ResNet ~\cite{he2016deep}, and GoogleNet ~\cite{szegedy2015going}, fall into this category. We demonstrate that in an L1LOS, its vulnerability is primarily governed by the linear operators. By exploiting their ill-conditioned characteristics, we can amplify the input perturbation by aligning it with the largest right singular vector of each linear operator.

In summary, our contributions can be summarized as follows:

1) We propose a novel data-free universal attack method called IntriUAP by exploiting the intrinsic vulnerability of a popular type of model called L1LOS which includes most deep models. 

2) We theoretically reveal that the vulnerability of an L1LOS model is majorly dominated by its linear parts, leading to the interpretability of the proposed IntriUAP.

3) Since our IntriUAP exploits the intrinsic vulnerability of the linear layers in deep models, it does not require full knowledge of the victim model to execute an attack. The IntriUAP can achieve a comparable attack success rate with only partial knowledge of the model's weights, making our method effective under a weak assumption of adversarial knowledge.

4) We benchmarked our method against the latest data-free and data-dependent UAPs on the ImageNet dataset. Our method achieves state-of-the-art performance among data-free approaches and is even comparable to some data-dependent methods in certain aspects.

\section{Related Work}

\textbf{Data-dependent Universal attack methods.} The pioneering work Universal Adversarial Perturbations \cite{moosavi2017universal} first revealed the existence of UAPs, generating them through iterative processes using DeepFool. SV-UAP method \cite{khrulkov2018art} employed the singular vectors of the Jacobian matrices of certain layers' feature maps to generate UAPs, achieving this with a minimal number of samples. Both Poursaeed et al. and Mopuri et al. introduced approaches to synthesize UAPs using generative models, termed GAP~\cite{poursaeed2018generative} and NAG~\cite{mopuri2018nag}, respectively. Zhang et al. proposed a technique for crafting UAPs aimed at specific target classes, dubbed CD-UAP~\cite{zhang2020cd}. Furthermore, Zhang's additional works, DF-UAP \cite{zhang2020understanding} and Cos-UAP \cite{zhang2021data}, leveraged dominant features and a cosine similarity loss function to create UAPs. Li et al. proposed an approach that integrates instance-specific and universal attacks from a feature perspective called AT-UAP~\cite{li2022learning}. Recently, Liu et al. introduced a novel technique employing stochastic gradient aggregation to create a more powerful UAP, termed SGA-UAP~\cite{liu2023enhancing}.

\textbf{Data-free Universal attack methods.} To generate powerful UAPs, dependency on samples still remains a pivotal issue. Currently, several data-free UAP methods have been developed. The earliest data-free approach, Fast Feature Fool~\cite{mopuri2017fast}, generates UAPs by maximizing activations at each layer. Subsequently, GD-UAP~\cite{mopuri2018generalizable} was introduced, enhancing FFF and extending its applicability to a broader range of vision tasks. Following this, the AAA~\cite{mopuri2018ask} method was proposed, employing class-impressions to create UAPs. PD-UA~\cite{liu2019universal}, utilized a Monte Carlo sampling approach to increase model uncertainty. Zhang et al. involved the use of proxy datasets for UAP generation, called AT-UAP~\cite{zhang2020understanding}, but this still relied on real samples. Later,  Zhang et al. introduced Cos-UAP~\cite{zhang2021data}, which completely replaced proxy datasets with manual samples, thus elevating the development of data-free UAPs to a new level.

\section{The Proposed Method} \label{sec:method}
In this section, we provide the motivation and formulate the optimization problem of our IntriUAP. We constrain our discussion to the classification task.  
\subsection{Problem Statement}
Consider a deep model classifier $f(\boldsymbol{x})$ with the input $\boldsymbol{x}$, the generation of a UAP $\boldsymbol{\xi}$ generally can be expressed as the following optimization problem:
\begin{equation}
\begin{gathered}
\underset{\boldsymbol{\xi}}{\arg \max } \frac{1}{n} \sum_{i=1}^{n} \mathcal{L}\left(f\left(\boldsymbol{x}_{i}+\boldsymbol{\xi}\right), \boldsymbol{y}_{i}\right), \text{s.t. } \|\boldsymbol{\xi}\|_{p} \leqslant \epsilon,
\end{gathered}
\end{equation}
where \( f\left(\boldsymbol{x}_{i} + \boldsymbol{\xi}\right) \) represents the predicted label, \(\boldsymbol{y}_{i}\) is the true label of \(\boldsymbol{x}_{i}\), and \(\mathcal{L}(\cdot)\) denotes the adversarial loss, such as cross-entropy. However, such problems have a flaw, namely that adversarial loss often requires samples and their true labels, which lack practicality due to data privacy and accessibility issues. 

To eliminate the assumption of access to the training or validation datasets, as well as the ground truth labels, we propose leveraging the intrinsic vulnerabilities of deep models to design UAP. More precisely, consider an $\ell$-layer neural network 
$f(\boldsymbol{x})$ 
described by the following recursive equations:
\begin{equation}
\begin{gathered}
f(\boldsymbol{x})=W_{\ell} \boldsymbol{x}_{\ell}+\boldsymbol{b}_{\ell}, \\
\boldsymbol{x}_{k+1}=\phi \left(W_{k} \boldsymbol{x}_{k}+\boldsymbol{b}_{k}\right), \quad \boldsymbol{x}_{1}=\boldsymbol{x},
\label{eq:03}
\end{gathered}
\end{equation}
where $k=1, \cdots, \ell-1$, $\boldsymbol{x}_{k}$ 
is the input feature of the $k$-th linear layer, 
$\phi$ represents a nonlinear layer, which can be an activation function, max pooling, etc., 
$W_{k}$ and $\boldsymbol{b}_{k}$ 
are the layer-wise weight matrix and bias vector respectively. Clearly, the above formula is an abstract representation of deep models, which consists of linear layers \footnote{In this paper, we collectively refer to linear and affine as linear, because the bias term would not introduce additional output error. } (such as convolutional layers, batch normalization layers, etc.) and nonlinear layers (activation functions, pooling layers, etc.). 

Now, suppose the model $f(\boldsymbol{x})$ belongs to the L1LOS category, i.e., the Lipschitz constant of all the nonlinear layers is equal to 1. We have the following important observations about  $f(\boldsymbol{x})$. \textbf{For an L1LOS model $f(\boldsymbol{x})$, its stability is dominated by its linear parts.} In other words, those nonlinear layers with Lipschitz constant 1 would not amplify the input perturbation to cause a large output error. Thus, we can focus on the linear parts and leverage their ill-conditioned characteristics to design UAPs. As for the linear part, we have another observation. \textbf{For a linear operator, the right singular vector corresponding to the maximal singular value could maximize the \(\ell_2\)-norm of the difference between the original and perturbed output of the linear operator}. The above two observations could guide the design of our proposed IntriUAP. That is, the ideal UAP should be aligned with the right singular vector corresponding to the maximal singular value. We will provide formal justification for the above two observations in Section \ref{sec:theory}. Before this, we first provide the detailed process of our proposed IntriUAP in the upcoming sections.

\subsection{The Formulation of Intrinsic UAP}
According to the discussion above, IntriUAP aims to design perturbations being aligned with the right singular vector corresponding to the maximal singular value of each linear layer of an L1LOS model. Specifically, let $\boldsymbol{\xi}$ be the UAP to be optimized. Obeying the notations in formula (\ref{eq:03}), the output error of the first linear layer caused by the input perturbation $\boldsymbol{\xi}$ is 
\begin{equation}
\boldsymbol{\delta}_{1+1}=W_{1}\left(\boldsymbol{x}_{1} + \boldsymbol{\xi} \right)-W_{1}\left(\boldsymbol{x}_{1}\right) = W_{1} \boldsymbol{\xi}.
\end{equation}
Similarly, using a recursive form, the output error of the $k$-th linear layer is as follows
\begin{equation}
\boldsymbol{\delta}_{k+1}=W_{k}\left(\boldsymbol{x}_{k}+\boldsymbol{\delta}_{k}\right)-W_{k}\left(\boldsymbol{x}_{x}\right) = W_{k}\boldsymbol{\delta}_{k}
\end{equation}
where $\boldsymbol{x}_{k}$ is the input of the $k$-th linear layer, and $\boldsymbol{\delta}_{k}$ is the perturbation transmitted from the front linear layers. 

Based on our second observation, for the $k$-th linear operation $\boldsymbol{\delta}_{k+1} = W_{k}\boldsymbol{\delta}_{k}$   ($k=1,...,\ell$), the right singular vector corresponding to the maximal singular value of $W_{k}$ would maximize the \(\ell_2\)-norm of the output of this linear layer (we denote this singular vector by $\boldsymbol{v}_{k}$). That is the optimal input perturbation $\boldsymbol{\delta}_{k}$ to maximize the $\|\boldsymbol{\delta}_{k+1}\|_2$ should be align with $\boldsymbol{v}_{k}$, i.e., $\boldsymbol{\delta}_{k}/\|\boldsymbol{\delta}_{k}\|_2 = \boldsymbol{v}_{k}$. By viewing the input perturbation $\boldsymbol{\delta}_{k}$ of each layer as a function of the expected UAP $\boldsymbol{\xi}$, the ideal UAP $\boldsymbol{\xi}$ that would cause the maximal model output error thus should be the one, making each $\boldsymbol{\delta}_{k}$ be align with $\boldsymbol{v}_{k}$. Therefore, the optimization problem of our IntriUAP could be formally expressed as follows  
\begin{equation}
\begin{gathered}
\underset{\boldsymbol{\xi}}{\arg \max } \sum_{k=1}^{\ell}\left|\left\langle\boldsymbol{\delta}_{k}(\boldsymbol{\xi}), \boldsymbol{v}_{k}\right\rangle\right| ,  \text { s.t. } \|\boldsymbol{\xi}\|_{p}  \leqslant \epsilon. 
\label{eq:opti}
\end{gathered}
\end{equation}
where the hyper-parameter $\epsilon$ controls the perturbation magnitude of the UAP $\boldsymbol{\xi}$.

We can observe several advantages of our IntriUAP from the optimization problem (\ref{eq:opti}). Firstly, this problem does not involve any input example $\boldsymbol{x}$ from the same distribution as the training dataset, which implies that IntriUAP is a data-free approach. Another notable aspect is that the problem does not require any training loss function, such as cross-entropy, which is commonly used in existing UPA methods. This indicates that solving the problem (\ref{eq:opti}) does not necessitate the use of the entire model; even a few linear layers can produce a satisfactory UAP. For instance, we can consider $\ell$ representing the first few linear layers in a model. This characteristic can partially limit the adversary's capabilities, making the attack scenario more realistic. We will validate these two advantages through our experiments in Section \ref{sec:exp}.

\subsection{The Algorithm for Intrinsic UAP}
In this section, we introduce the details of generating Intrinsic UAP. We follow the optimization procedure outlined in Algorithm~\ref{alg:algorithm}. Firstly, we focus on an L1LOS deep model, which is inherently compatible with our intrinsic attack methodology. We then calculate all of its maximum singular vectors \(\boldsymbol{v}_{k}\) for \(k = 1, \dots, \ell \). The initial data $\boldsymbol{x}$ can be a range prior, Gaussian noise, or uniform noise. We employ the Adam optimizer combined with a StepLR scheduler.

Next, during the optimization phase of Algorithm~\ref{alg:algorithm}, lines 3 and 4 compute the perturbed outputs \(\boldsymbol{\delta_{k}}\) for each linear layer. Lines 6 to 8 involve calculating the inner product of \(\boldsymbol{\delta_{k}}\) with the corresponding singular vectors \(\boldsymbol{v}_{k}\), performing backpropagation, and optimizing the initial perturbation \(\boldsymbol{\xi}\). Finally, a clipping operation is applied to ensure that the \(\ell_\infty\)-norm of \(\boldsymbol{\xi}\) remains within the bound of 10. The above process is repeated until the loss converges.


\begin{algorithm}[t]
\caption{Algorithm for Our Proposed IntriUAP Method}
\label{alg:algorithm}
\begin{algorithmic}[1] 

\REQUIRE Target $l$-layers CNN $f$, initial data $\boldsymbol{x}$, initial perturbation $\boldsymbol{\xi}$, Scheduler, learning rate $\gamma$, maximum perturbation value $\ell_{\infty}$-norm $\epsilon$, epoch number $T$.
\ENSURE Perturbation $\boldsymbol{\xi}$

\STATE Calculate the maximum singular vectors $\boldsymbol{v}_{k}$ for $k = 1, \dots, \ell$ of all linear layers of $f$.

\FOR{$t = 1$ to $T$}
    \FOR{$k = 1$ to $\ell$}
        \STATE $\boldsymbol{\delta}_{k+1} \gets W_{k}(\boldsymbol{x}_{k} + \boldsymbol{\delta}_{k}) - W_{k}(\boldsymbol{x}_{k})$
    \ENDFOR
    \STATE $\mathcal{L} \gets -\sum_{k=1}^{\ell} |\langle \boldsymbol{\delta}_{k}(\boldsymbol{\xi}), \boldsymbol{v}_{k} \rangle|$
    \STATE \textbf{Backpropagate} the loss $\mathcal{L}$ to compute gradients $\nabla_{\boldsymbol{\xi}} \mathcal{L}$
    \STATE \textbf{Update} the perturbation $\boldsymbol{\xi}$ using gradient descent: $\boldsymbol{\xi} \gets \boldsymbol{\xi} - \gamma \nabla_{\boldsymbol{\xi}} \mathcal{L}$
    \STATE $\boldsymbol{\xi} \gets \text{Clip}^{\epsilon}(\boldsymbol{\xi})$
\ENDFOR

\RETURN $\boldsymbol{\xi}$

\end{algorithmic}
\end{algorithm}

\section{Deep Models as L1LOS} \label{sec:l1los}
In this section, we will demonstrate that certain popular networks, including VGG, ResNet, GoogleNet, AlexNet, etc., are L1LOS. As a result, our proposed UAP is fully compatible with the task of attacking these widely used deep models. Since our proposed method relies on the singular vectors of the linear layers, we then discuss how to represent the widely adopted linear layers, the convolutional layers, and the BatchNorm layers as matrix operations. 

\subsection{View Popular Deep Models as L1LOS}
For networks such as VGG, GoogleNet, and ResNet, the linear components during the inference stage consist of convolutional layers, batch normalization layers, and fully connected layers. The nonlinear components are primarily composed of ReLU activations and max-pooling layers, both of which have a Lipschitz constant of 1 \cite{kim2021lipschitz}.

For networks like VGG, the L1LOS structure is naturally satisfied. In the case of GoogleNet with its Inception modules, each branch within an Inception module adheres to the L1LOS structure, with a final concatenate operation. For ResNet, with its residual block, Assuming a residual block $\mathcal{F}$ consists of two convolutional layers \( W_1 \) and \( W_2 \), along with corresponding normalization layers \( \operatorname{BN}_1 \) and \( \operatorname{BN}_2 \), as well as their respective nonlinear activations $\sigma$, the structure of the residual block can be expressed as:
\begin{equation}
\mathcal{F}(\boldsymbol{x}) = \sigma(\operatorname{BN}_2(W_2 \ast \sigma(\operatorname{BN}_1(W_1 \ast \boldsymbol{x}))))+\boldsymbol{x}.
\end{equation}
Although connecting multiple residual blocks introduces more complex nonlinearity, ResNet still satisfies the L1LOS structure.

\subsection{Convolutional Layers as a Linear Operator}
There are already existing works that represent convolutional kernels as matrices in different ways~\cite{sedghi2018singular,araujo2021lipschitz,praggastis2022svd}. We know that a 2-D convolutional kernel can be represented as a doubly-block Toeplitz matrix.

Given any kernel coefficients $K$, the matrix representation for the convolution by $K$ is represented by the following doubly-block circulant matrix:
\begin{equation}
W = \begin{bmatrix}
\text{circ}(K_{0,:}) & \text{circ}(K_{1,:}) & \cdots & \text{circ}(K_{n-1,:}) \\
\text{circ}(K_{1,:}) & \text{circ}(K_{2,:}) & \cdots & \text{circ}(K_{0,:}) \\
\vdots & \vdots & \ddots & \vdots \\
\text{circ}(K_{n-1,:}) & \text{circ}(K_{0,:}) & \cdots & \text{circ}(K_{n-2,:})
\end{bmatrix}
\end{equation}
where \(\text{circ}(K_{i,:})\) is the circulant matrix formed from the \(i\)-th row of the matrix \(K\), where each row of the circulant matrix is a right cyclic shift of the previous row, and $W$ is the Toeplitz matrix representing the convolution \cite[Page~329]{Goodfellow-et-al-2016}). By vectorizing the input matrix $X$, we can represent the convolution operation as the matrix multiplication $\textnormal{vec}(Y) = W \cdot \textnormal{vec}(X)$. More details for gain $W$ and presenting a convolutional layer as the matrix operation can be found in the supplementary material. 

\subsection{BatchNorm Layer as a Linear Operator}

The BatchNorm layer is widely used in CNNs to accelerate training and enhance stability by normalizing the input features. During the training phase, BatchNorm includes nonlinear operations, such as the calculation of batch statistics. However, during inference, when batch statistics are replaced by moving averages, the BatchNorm layer can be simplified to a linear transformation. Since the task of UAP is to attack a well-trained model, therefore it is reasonable to treat the BatchNorm layer as a linear operator. Then, we introduce how to represent it as a matrix operation. 

Given an input \(\mathbf{X} \in \mathbb{R}^{c \times m \times m}\) to the BatchNorm layer, where \(c\) is the number of channels and \(m \times m\) is the spatial dimension, the output \(\mathbf{Y}\) can be expressed as \(\textnormal{vec}(\mathbf{Y}) = A \cdot \textnormal{vec}(\mathbf{X}) + \boldsymbol{b}\). Here, \(A\) is a diagonal matrix in \(\mathbb{R}^{(c \times m \times m)\times(c \times m \times m)}\), which represents the linear transformation, while \(\boldsymbol{b}\) accounts for the shift. The diagonal elements of \(A\), denoted as \(A_{ii}\), are defined as:
\begin{equation}
	A_{ii} = \frac{\gamma_{c}}{\sqrt{v_{c} + \varepsilon}}, \quad c = \left\lfloor \frac{i}{m \times m} \right\rfloor + 1,
\end{equation}
where \(\gamma_c\) is the scale factor, \(v_c\) is the moving variance for channel \(c\), \(\varepsilon\) is a small constant added for numerical stability, and \(\left\lfloor \cdot \right\rfloor\) denotes the floor function. The index \(i\) ranges over all elements in the input tensor, \(i \in [0, 1, \ldots, c \times m \times m - 1]\). The vector \(\boldsymbol{b} \in \mathbb{R}^{(c \times m \times m) \times 1}\) is the bias term, with each element \(\boldsymbol{b}_{i}\) given by:
\begin{equation}
\boldsymbol{b}_{i} = \beta_{c} - \mu_{c} \times \frac{\gamma_{c}}{\sqrt{v_{c} + \varepsilon}},
\end{equation}
where \(\beta_c\) is the shift factor and \(\mu_c\) is the moving mean for channel \(c\).  

\section{Theoretical Justification}
\label{sec:theory}
 In this section, we provide the theoretical justification of the two observations inspiring our IntriUAP which we have briefly introduced in Section \ref{sec:method}. The detailed proof can be found in the supplementary material. The first observation is that \textbf{for a system composed of nonlinear operators with the Lipschitz constant is 1 and linear operators, (i.e., L1LOS), the stability of this system is dominated by its linear parts.} The formal justification of this observation is as follows:

\newtheorem{theorem}{Theorem}
\begin{theorem}
For a L1LOS $f(\boldsymbol{x})$ defined in formula (\ref{eq:03}), we have $\operatorname{Lip}\left(f\right) \leq \prod_{k=0}^{\ell-1}\left\|W_{k}\right\|_{op},$ 
where $\operatorname{Lip}\left(f\right)$ is the Lipschitz constant of $f(\boldsymbol{x})$.
\label{theo1}
\end{theorem}

As indicated by the above theorem, the upper bound on the Lipschitz constant for a model $f$ is primarily governed by its linear operators. This implies that integrating 1-Lipschitz nonlinear operators into a linear system will not increase the Lipschitz constant of this model. It is easy to show that 1-Lipschitz nonlinear operators will not amplify the perturbation according to the definition of the Lipschitz constant. Thus, we only need to consider linear operators for amplifying the input perturbation. 

Upon having the above conclusion, we also need to know what kind of input perturbation would be amplified by the linear operators to cause a large output error. We thus have the second observation discussed in Section \ref{sec:method}. 2) \textbf{For a linear operator, the right singular vector corresponding to the maximal singular value could maximize the output of this operator measured by the $L_p$-norm}, which can be supported by the following theorem 

\begin{theorem} 
For any bounded linear operator, the operator norm \( \|A\|_{op} \) is equal to its largest singular value \( \sigma_{\max}(A) \), and the corresponding right singular vector \( \boldsymbol{v}_{\max} \) achieves this maximum, such that:
\[
\|A\|_{op} = \sigma_{\max}(A) = \sup_{\|\boldsymbol{x}\|_2=1} \|A\boldsymbol{x}\|_2 = \|A\boldsymbol{v}_{\max}\|_2,
\]
where \( \boldsymbol{v}_{\max} \) is the right singular vector corresponding to \( \sigma_{\max}(A) \).
		\label{lema:4}
\label{Cor:1}
\end{theorem}

This theorem shows that the unit vector to cause the maximal output value under the \(\ell_2\)-norm is $\boldsymbol{v}_{\max}$. Therefore, if an input perturbation $\boldsymbol{\xi}$ with a given magnitude $\epsilon$ wants to maximize the output value of a linear operator, it should be align with the vector $\boldsymbol{v}_{\max}$, i.e., $\boldsymbol{\xi} = \epsilon \boldsymbol{v}_{\max}$.   

\section{Experiment} \label{sec:exp}
In this section, we present the experimental setup and results to evaluate the effectiveness of our proposed method and our claims. First, we assess the method’s effectiveness in the straightforward white-box scenario. Second, to evaluate transferability, we conduct experiments in a black-box setting. Third, we explore the robustness of the method against some image preprocessing methods. Finally, due to the special properties of our approach, we also perform semi-white-box experiments, i.e., only access a few layers in a model.

\subsection{Experimental Setup}
\noindent \textbf{Initialization of IntriUAP.} We consider the following initialization of our IntriUAP $\boldsymbol{\xi}$ describe in Algorithm \ref{alg:algorithm}:
\begin{enumerate}
	\item \textbf{ImageNet Mean and Range prior:} Inputs were generated by leveraging the ImageNet mean prior and give it a dynamic range. This involved creating a three-channel RGB image with channel-wise values set to [0.485, 0.456, 0.406] and each elements are then given a dynamic range, aligning with the Range prior introduced in the seminal work of Mopuri et al. \cite{mopuri2018generalizable}.
	
	\item \textbf{Gaussian Distribution:} We generated perturbations by sampling from a Gaussian distribution \(\mathcal{N}(\mu, \sigma^2)\). In our experiments, \(\mu\) was set to 0.45, with \(\sigma\) values of 0.1.

	\item \textbf{Uniform Distribution:} We generated perturbations by sampling from a uniform distribution \(\mathcal{U}(a, b)\). In our experiments, $a$ was set to 0.40, $b$ was set to 0.60.

\end{enumerate}

\noindent \textbf{Victim Models.} We consider five popular classification models provided by torchvision, including AlexNet~\cite{krizhevsky2012imagenet}, GoogleNet~\cite{szegedy2015going}, VGG-16~\cite{simonyan2014very}, VGG-19~\cite{simonyan2014very} and ResNet152~\cite{he2016deep}. Note that all the considered models are L1LOS as we discussed in Section \ref{sec:l1los}.

\noindent \textbf{Evaluation metrics.} To effectively assess the attack performance of our method, we report the fooling ratio on the 50,000-image validation set from ImageNet ILSVRC2012, a metric widely used in UAP tasks \cite{moosavi2017universal,poursaeed2018generative,mopuri2018nag,zhang2020understanding,zhang2021data,li2022learning,liu2023enhancing}. The fooling ratio is obtained by calculating the proportion of samples with labels changes when applying UAP. \(\ell_\infty\)-norm of the UAP remains within the bound of 10 (rescaled to [0,255]).

\noindent \textbf{Baselines.} The proposed method is compared with the following UAP methods in the white-box and black-box attack scenario:
\begin{enumerate}

\item \textbf{Data-dependent UAP methods:} UAP~\cite{moosavi2017universal}, SV-UAP~\cite{khrulkov2018art}, NAG~\cite{mopuri2018nag}, GAP~\cite{poursaeed2018generative}, DF-UAP~\cite{zhang2020understanding}, Cos-UAP~\cite{zhang2021data}, AT-UAP~\cite{li2022learning}, SPGD-UAP~\cite{liu2023enhancing}.

\item \textbf{Data-free UAP methods:} FFF-UAP~\cite{mopuri2017fast}, AAA-UAP~\cite{mopuri2018ask}, GD-UAP~\cite{mopuri2018generalizable}, PD-UAP~\cite{liu2019universal}, DF-UAP~\cite{zhang2020understanding}, Cos-UAP~\cite{zhang2021data}.
\end{enumerate}

\begin{table}[t]
    \centering
    \fontsize{8.5}{9}\selectfont
    \setlength{\tabcolsep}{3pt}
    \begin{tabular}{lccccc}
        \toprule
        Method    & AlexNet & GoogleNet & VGG16 & VGG19 & ResNet152 \\
        \midrule
        UAP       & 93.30   & 79.90     & 78.30   & 77.80   & 84.00   \\
        SV-UAP    & -- --   & -- --     & 52.00   & 60.00   & -- --   \\
        GAP       & -- --   & 82.70     & 83.70   & 80.10   & -- --   \\
        NAG       & 96.44   & 90.37     & 77.57   & 83.78   & 87.24   \\
        AT-UAP    & 97.01   & 90.82     & 97.51   & 97.56   & 91.52   \\
        SGA-UAP   & 97.43   & 92.12     & 98.36   & 97.69   & 94.04   \\
        \midrule
        FFF       & 80.92   & 56.44     & 47.10   & 43.62   & -- --   \\
        AAA       & 89.04   & 75.28     & 71.59   & 72.84   & 60.72   \\
        GD-UAP    & 87.02   & 71.44     & 63.08   & 64.67   & 37.30   \\
        PD-UAP    & -- --   & 67.12     & 53.09   & 48.95   & 53.51   \\
        DF-UAP    & 89.90   & 76.80     & 92.20   & 91.60   & \textbf{79.9} \\
        Cos-UAP   & 91.07   & \textbf{87.57} & 89.48   & 86.81   & 65.35   \\
        Ours      & \textbf{94.03} & 60.42 & \textbf{93.40} & \textbf{92.28} & 65.47   \\
        \bottomrule
    \end{tabular}
    \caption{Fooling ratio (\%) of different UAP generation methods in the white-box attack scenario. The results are divided into universal attacks with access to the original ImageNet training data (upper) and data-free methods (lower).}
    \label{label1}
\end{table}

\subsection{Effectiveness of Intrinsic UAPs}

Table~\ref{label1} provides a summary of nearly all state-of-the-art data-free and data-dependent methods. We compare our Intri-UAP approach with \cite{mopuri2017fast, mopuri2018ask, mopuri2018generalizable, liu2019universal, zhang2020understanding, zhang2021data}, as they don't rely on real training samples. In the experimental results table, we use $"--"$ to indicate items for which the other methods did not conduct experiments. As can be seen from Table~\ref{label1}, our method outperforms the competitive data-free methods in most cases. Even for the comparison of the data-driven methods (the upper half table), we still achieve comparable performance. Note, that the DF-UAP method utilizes some images from the COCO dataset. 

A comparison of IntriUAP with these existing strictly data-free methods is presented in Table~\ref{ddsd}. Our results demonstrate that UAPs generated on AlexNet and VGG achieve a high fooling ratio, showing less dependency on the sample data compared to other models. Additionally, ResNet also delivers commendable results.

\begin{table}[!t]
    \centering
    \fontsize{8}{8.5}\selectfont
    \setlength{\tabcolsep}{1.5pt} 
    \begin{tabular}{c c c c c c c}
        \toprule
        Initial Data & Method & AlexNet & GoogleNet & VGG16 & VGG19 & ResNet152 \\ 
        \midrule
        \multirow{3}{*}{Range Prior} 
        & PD-UAP & -- -- & -- -- & 70.69 & 64.98 & 46.39 \\ 
        & GD-UAP & 87.02 & \textbf{71.44} & 63.08 & 64.67 & 37.30 \\ 
        & Ours & \textbf{91.60} & 60.42 & \textbf{93.40} & \textbf{92.14} & \textbf{65.47} \\ 
        \midrule
        \multirow{3}{*}{Uniform} 
        & Cos-UAP & 82.60 & 40.30 & 72.30 & 64.40 & 47.20 \\ 
        & Ours & \textbf{92.58} & \textbf{45.56} & \textbf{80.40} & \textbf{81.56} & \textbf{48.00} \\ 
        \midrule
        \multirow{3}{*}{Gaussian}
        & AT-UAP & 56.80 & 24.27 & 30.56 & 27.75 & 19.40 \\ 
        & Cos-UAP & 89.50 & 46.70 & 76.10 & 75.40 & 49.90 \\ 
        & Ours & \textbf{92.61} & \textbf{47.80} & \textbf{86.03} & \textbf{82.00} & \textbf{50.50} \\ 
        \bottomrule
    \end{tabular}
    \caption{Results for different initialization methods of our IntriUAP in comparison with different data-free methods.}
    \label{ddsd}
\end{table}

	\begin{figure*}[ht]
    \centering
    \begin{minipage}[b]{0.15\textwidth}
        \centering
        \includegraphics[width=\linewidth]{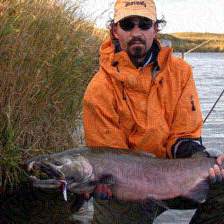}
        \caption*{Jigsaw}
        \label{fig:image1}
    \end{minipage}
    \hfill
    \begin{minipage}[b]{0.15\textwidth}
        \centering
        \includegraphics[width=\linewidth]{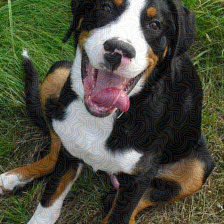}
        \caption*{Brain coral}
        \label{fig:image2}
    \end{minipage}
    \hfill
    \begin{minipage}[b]{0.15\textwidth}
        \centering
        \includegraphics[width=\linewidth]{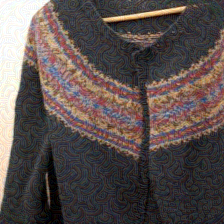}
        \caption*{Brain coral}
        \label{fig:image3}
    \end{minipage}
    \hfill
    \begin{minipage}[b]{0.15\textwidth}
        \centering
        \includegraphics[width=\linewidth]{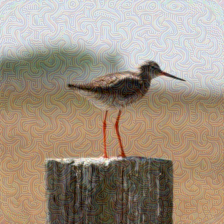}
        \caption*{Brain coral}
        \label{fig:image4}
    \end{minipage}
    \hfill
    \begin{minipage}[b]{0.15\textwidth}
        \centering
        \includegraphics[width=\linewidth]{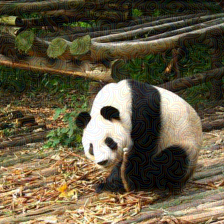}
        \caption*{Brain coral}
        \label{fig:image5}
    \end{minipage}
    \hfill
    \begin{minipage}[b]{0.15\textwidth}
        \centering
        \includegraphics[width=\linewidth]{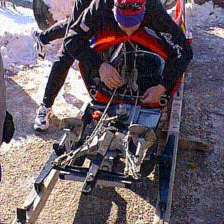}
        \caption*{Taxi}
        \label{fig:image6}
    \end{minipage}
    
    \caption{Examples of perturbed images generated by VGG19 with IntriUAP and their corresponding labels. The perturbations were constrained with $\ell_{\infty} \text {-norm } = 10 $. Using range prior.}
    \label{fig:images}
\end{figure*}

In Table \ref{your_label123123}, we compared our approach with the current state-of-the-art data-dependent methods, SPGD and SGA\cite{liu2023enhancing}. Remarkably, even when they utilize 500 real samples, our method still holds an advantage. For example, we achieve that fooling rate of $93.40\%$ on attacking VGG16 while the best competitor is $89.83\%$

We also provide the visualization of our IntriUAP in Figure \ref{fig:images}. As can be seen, the IntriUAPs have not significantly impacted the quality of the cover images. 
\begin{table}[!t]
    \centering
    \fontsize{8}{8.5}\selectfont
    \setlength{\tabcolsep}{1.0pt} 
    \begin{tabular}{lcccccc}
        \toprule
        Method  & Samples & AlexNet & GoogleNet & VGG16 & VGG19 & ResNet152 \\
        \midrule
        Singular Fool* & 49 & 52.00 & 60.00 & 44.00 & -- -- & -- -- \\
        GD-UAP* & 49 & 72.80 & 67.60 & 56.40 & -- -- & -- -- \\
        UAP* & 500 & 57.33 & 16.61 & 25.29 & 25.04 & 19.11 \\
        GAP* & 500 & 86.89 & 57.07 & 70.40 & 65.89 & 47.58 \\
        SPGD* & 500 & 92.35 & 41.68 & 81.70 & 75.74 & 23.44 \\
        SGA* & 500 & 93.03 & 68.33 & 89.83 & 88.70 & 52.12 \\
        Cos-UAP & 0 & 91.07 & \textbf{87.57} & 89.48 & 86.81 & 65.35 \\		
        Ours & \textbf{0} & \textbf{94.03} & 60.43 & \textbf{93.40} & \textbf{92.28} & \textbf{65.47} \\
        \bottomrule
    \end{tabular}
    \caption{Fooling ratio (\%) for UAPs crafted with limited real samples. * indicates the data-dependent method.} 
    \label{your_label123123}
\end{table}

\subsection{Transferability Performance of Intrinsic UAPs}
In this section, the transferability results are presented in Table~\ref{tab:uap_trans}. Each row in the table shows the fooling rates for perturbations learned on a specific target model when attacking various other models (columns). These ratios are derived using the IntriUAP objective with range prior. Diagonal values correspond to white-box attack scenarios, while off-diagonal values represent black-box attacks. Our observations indicate that the transferability of our UAPs is effective, particularly when conducting black-box attacks using UAPs generated with L1LOS.

In Table~\ref{tab:uap_transfer}, we also present the results of our experiments in black-box scenarios, compared with both data-free and data-dependent approaches, for evaluation of the transferability of those UAPS. Notably, even when compared to data-dependent methods, our approach demonstrates superior performance in black-box scenarios, even surpassing the current data-dependent method SGA in black-box attacks on AlexNet. Furthermore, we found that VGG and AlexNet often exhibit good transferability among themselves. UAPs generated by Intrinsic Attack trained on AlexNet demonstrate the strongest transferability.

\begin{table}[!t]
    \centering
    \fontsize{9}{9.5}\selectfont
    \setlength{\tabcolsep}{2.5pt} 
    \begin{tabular}{lccccc}
        \toprule
        Model & AlexNet & GoogleNet & VGG16 & VGG19 & ResNet152 \\
        \midrule
        AlexNet & \textbf{94.07} & 56.79 & 71.14 & 65.74 & 41.11 \\
        GoogleNet & 41.44 & \textbf{60.20} & 48.40 & 45.19 & 28.17 \\
        VGG16 & 43.39 & 36.80 & \textbf{93.40} & 82.18 & 35.67 \\
        VGG19 & 42.03 & 39.46 & 85.69 & \textbf{92.28} & 39.96 \\
        ResNet152 & 42.41 & 38.45 & 53.05 & 51.56 & \textbf{65.47} \\
        \bottomrule
    \end{tabular}
    \caption{Fooling ratio (\%) of UAPs under Black-box settings.}
    \label{tab:uap_trans}
\end{table}

\begin{table}[!t]
    \centering
    \fontsize{7.5}{8}\selectfont
    \setlength{\tabcolsep}{1.5pt} 
    \begin{tabular}{lrrrrrrrrrr}
        \toprule
        Model & \multicolumn{2}{c}{AlexNet} & \multicolumn{2}{c}{GoogLeNet} & \multicolumn{2}{c}{VGG-16} & \multicolumn{2}{c}{VGG-19} & \multicolumn{2}{c}{ResNet-152} \\
        \cmidrule(lr){2-3} \cmidrule(lr){4-5} \cmidrule(lr){6-7} \cmidrule(lr){8-9} \cmidrule(lr){10-11}
        Method & GD & Ours & GD & Ours & GD & Ours & GD & Ours & GD & Ours \\
        \midrule
        No Defense & 87.10 & \textbf{94.07} & \textbf{71.40} & 60.42 & 63.10 & \textbf{93.40} & 64.70 & \textbf{92.28} & 37.30 & \textbf{65.47} \\
        \midrule
        75\% JPEG & 72.10 & \textbf{90.40} & 41.80 & \textbf{44.51} & 49.12 & \textbf{61.19} & \textbf{64.70} & 55.47 & 37.30 & \textbf{38.30} \\
        50\% JPEG & 79.20 & \textbf{85.88} & 37.85 & \textbf{42.54} & 39.12 & \textbf{49.40} & 54.07 & \textbf{54.93} & 25.84 & \textbf{33.23} \\
        \midrule
        Gaussian & 65.54 & \textbf{72.17} & 35.62 & \textbf{46.31} & 37.77 & \textbf{51.48} & 35.54 & \textbf{52.80} & 24.56 & \textbf{32.06} \\
        Median & 72.13 & \textbf{83.98} & 46.98 & \textbf{54.47} & 44.68 & \textbf{78.13} & 46.67 & \textbf{77.23} & 28.82 & \textbf{41.88} \\
        Bilateral & 34.80 & \textbf{67.20} & 21.50 & \textbf{26.74} & 35.80 & \textbf{70.52} & 28.20 & \textbf{66.78} & 25.40 & \textbf{35.07} \\
        \bottomrule
    \end{tabular}
    \caption{Fooling rate (\%) of UAPs under different defenses.}
    \label{tab:r}
\end{table}

\begin{table}[!t]
    \centering
    \fontsize{7.5}{8}\selectfont
    \setlength{\tabcolsep}{1.5pt} 
    \begin{tabular}{lcccc}
        \toprule
        Model & 25$\%$ white-box & 50$\%$ white-box & 75$\%$ white-box & 100$\%$ white-box \\
        \midrule
        AlexNet & 81.35 & 90.01 & 92.77 & \textbf{94.03} \\
        VGG16 & 57.95 & 67.10 & 91.49 & \textbf{93.40} \\
        VGG19 & 73.51 & 78.62 & 91.31 & \textbf{92.28} \\
        GoogleNet & 40.65 & 56.20 & 59.81 & \textbf{60.43} \\
        \bottomrule
    \end{tabular}
    \caption{Fooling ratio ($\%$) of UAPs for semi-white-box model. Generated with range prior as background. The rows indicate the white-box level of the model, and the columns indicate the target model.}
    \label{tab:ff_ddd}
\end{table}

\begin{table*}[!t]
    \centering
    \fontsize{9}{10}\selectfont
    \begin{tabular}{|c|c|c|c|c|c|c|c|}
        \hline
        Target Model & Technique & AlexNet & VGG16 & VGG19 & ResNet50 & ResNet152 & GoogleNet \\ \hline
        \multirow{10}{*}{VGG16} 
        & UAP* & 33.35 & 76.73 & 64.14 & -- -- & 29.39 & 33.51 \\ \cline{2-8}
        & GAP* & 22.33 & 82.21 & 76.30 & -- -- & 29.46 & 42.50 \\ \cline{2-8}
        & AAA & -- -- & 71.59 & 65.64 & -- -- & 45.33 & \textbf{60.74} \\ \cline{2-8}
        & GD-UAP & -- -- & 45.47 & 38.20 & 27.70 & 23.80 & 34.13 \\ \cline{2-8}
        & GD-UAP+P & -- -- & 51.63 & 44.07 & 32.23 & 28.78 & 36.79 \\ \cline{2-8}
        & UA & -- -- & 48.46 & 41.97 & 29.09 & 24.90 & 35.52 \\ \cline{2-8}
        & PD-UA & -- -- & 53.09 & 49.30 & 33.61 & 30.31 & 39.05 \\ \cline{2-8}
        & Cos-UAP (Regular UAP) & -- -- & 89.48 & 76.84 & 44.11 & 38.37 & 48.97 \\ \cline{2-8}
        & Cos-UAP (Adaptive UAP) & -- -- & 86.16 & 77.88 & 49.30 & \textbf{44.27} & 56.96 \\ \cline{2-8}
        & Ours (Range prior) & \textbf{43.39} & \textbf{93.40} & \textbf{82.18} & \textbf{50.28} & 35.67 & 35.90 \\ \hline
        
        \multirow{7}{*}{VGG19} 
        & UAP* & 34.45 & 65.46 & 77.79 & 33.24 & 28.49 & 35.21 \\ \cline{2-8}
        & GAP* & 52.71 & 75.08 & 79.11 & 39.43 & 35.21 & \textbf{49.11} \\ \cline{2-8}
        & FFF & 42.03 & 38.19 & 43.62 & 28.27 & 26.34 & 30.71 \\ \cline{2-8}
        & GD-UAP & -- -- & \textbf{55.70} & 64.70 & -- -- & 35.80 & 53.50 \\ \cline{2-8}
        & UAP-DL & -- -- & 47.50 & 52.00 & -- -- & 30.40 & 33.70 \\ \cline{2-8}
        & DF-UAP (COCO) & -- -- & 83.40 & 91.70 & -- -- & 35.40 & 39.80 \\ \cline{2-8}
        & Ours (Range prior) & 42.80 & \textbf{85.60} & \textbf{92.28} & \textbf{55.87} & \textbf{38.93} & 38.65 \\ \hline
        
        \multirow{8}{*}{AlexNet} 
        & UAP* & 86.53 & 37.67 & 35.47 & 23.45 & 20.99 & 27.82 \\ \cline{2-8}
        & GAP* & 89.06 & 52.02 & 48.60 & 42.54 & 38.70 & 33.05 \\ \cline{2-8}
        & SGA-logit* & 95.23 & 57.62 & 53.86 & -- -- & 30.39 & 37.68 \\ \cline{2-8}
        & SPGD-logit* & 96.60 & 63.82 & 59.52 & -- -- & 34.95 & 46.18 \\ \cline{2-8}
        & SPGD-cls* & 95.97 & 58.59 & 54.03 & -- -- & 30.52 & 42.78 \\ \cline{2-8}
        & SGA-cls* & \textbf{97.23} & 66.46 & 60.60 & -- -- & 35.29 & 48.97 \\ \cline{2-8}
        & GD-UAP (Range Prior) & 87.02 & 50.46 & 49.92 & -- -- & 38.58 & 49.40 \\ \cline{2-8}
        & DF-UAP (COCO) & 90.45 & 60.43 & 58.66 & -- -- & \textbf{47.02} & 54.77 \\ \cline{2-8}
        & Ours (Range prior) & 94.07 & \textbf{71.14} & \textbf{65.74} & \textbf{50.07} & 41.11 & \textbf{56.79} \\ \hline
    \end{tabular}
    \caption{The fooling ratio~(\%) on six models in the black-box setting by regular UAP attack methods. The UAPs are crafted on AlexNet, VGG16, and VGG19 respectively. * indicates the data-dependent method.}
    \label{tab:uap_transfer}
\end{table*}

\subsection{Robustness of Intrinsic UAPs}

Following GD-UAP's approach, we evaluated UAPs against defensive techniques, focusing on input transformations such as JPEG compression, median smoothing, bilateral smoothing, and Gaussian blurring. Results in Table~\ref{tab:r} show our method's robustness significantly surpasses GD-UAP. 

The findings closely align with the conclusions reached by GD-UAP \cite{mopuri2018generalizable}. While input transformations do reduce the fooling rates of UAPs, they also degrade image quality and significantly lower the model's Top-1 accuracy. This reduction in accuracy is often unacceptable, highlighting the limited adaptability of these defense mechanisms.

\subsection{Intrinsic UAPs for Semi-white-box models}
Due to the special property of our methond, specifically that our method does not require backpropagation through the entire network, we conducted experiments to demonstrate the feasibility of attacking semi-white-box models. We performed experiments by selecting subsets of linear layers and categorized the degree of white-box access into four scenarios: 25\% white-box, 50\% white-box, 75\% white-box, and 100\% white-box (for example, 50\% white-box indicates access to only the first 50\% of the model's parameters). We observed that the fooling ratio remained high across these scenarios. The results indicate that partial access to the model's parameters still allows for the creation of effective UAPs. 

Table~\ref{tab:ff_ddd} presents the experimental results using VGG16, VGG19, GoogLeNet, and AlexNet. The results show that even with partial white-box access, the success rate remains high. For instance, AlexNet achieves 90\% fooling ratio at 50\% white-box access and only improves slightly to 94.03\% at 100\% access. Similarly, VGG16 reaches 91\% at 75\% white-box access and 93\% at 100\%, with minimal gains between 75\% and 100\%. This indicates that semi-white-box scenarios (50\% and 75\% access) can achieve nearly the same high success rate as full white-box access.

\section{Conclusion}

 In this paper, we explore the intrinsic vulnerability of the popular deep models called L1LOS, which consists of linear layers and 1-Lipschitz nonlinear layers. We point out that the vulnerability of L1LOS is controlled by linear operators. This deepens the theoretical foundation for the existence of UAPs and introduces the method of the Intrinsic UAP. We demonstrate that the vulnerability of L1LOS can be exploited without requiring samples or complete white-box access, achieving high attack success rates and posing the security risks. For L1LOS, which align with our theoretical framework, our fooling ratio achieves SOTA baselines. Compared with the UAP methods, our method not only has the weak assumption of data-free but also works under the weak assumption of access a few layers of the victim models, making our method more practical in real word tasks.

\section*{Acknowledgments}
This research was partially supported by the National Natural Science Foundation of China (Grant No. 62202009), the Macau Science and Technology Development Fund (Grant Nos. 0040/2023/ITP1 and 0004/2023/RIB1), and the Basic and Applied Basic Research Foundation of Guangdong Province (Grant No. 2024A1515011755).


\end{document}